\documentclass[letterpaper]{article} 
\usepackage{aaai23}  
\usepackage{times}  
\usepackage{helvet}  
\usepackage{courier}  
\usepackage[hyphens]{url}  
\usepackage{graphicx} 
\usepackage{natbib}  
\usepackage{caption} 
\usepackage{algorithm}
\usepackage{algorithmic}

\usepackage[caption=false]{subfig}
\usepackage{multirow}
\usepackage{amsmath}
\usepackage{verbatimbox}
\usepackage{amsfonts}
\usepackage{xcolor}
\usepackage[switch]{lineno}

\usepackage{newfloat}
\usepackage{listings}
\DeclareCaptionStyle{ruled}{labelfont=normalfont,labelsep=colon,strut=off} 
\floatstyle{ruled}
\newfloat{listing}{tb}{lst}{}
\floatname{listing}{Listing}
\title{FlowFace: Semantic Flow-guided Shape-aware Face Swapping}
\author{
    Hao Zeng\textsuperscript{\rm 1},
    Wei Zhang\textsuperscript{\rm 1},
    Changjie Fan\textsuperscript{\rm 1},
    Tangjie Lv\textsuperscript{\rm 1},
    Suzhen Wang\textsuperscript{\rm 1},\\
    Zhimeng Zhang\textsuperscript{\rm 1},
    Bowen Ma\textsuperscript{\rm 1},
    Lincheng Li\textsuperscript{\rm 1},
    Yu Ding\textsuperscript{\rm 1,}\textsuperscript{\rm 3,}\thanks{Yu Ding is the corresponding author.},
    Xin Yu\textsuperscript{\rm 2}
}
\affiliations{
    \textsuperscript{\rm 1}Virtual Human Group, Netease Fuxi AI Lab\\
    \textsuperscript{\rm 2}University of Technology Sydney\\
    \textsuperscript{\rm 3}Zhejiang University

    zenghao1110@gmail.com,
    \{zhengwei05,
    fanchangjie,
    hzlvtangjie,
    wangsuzhen,\\
    zhangzhimeng,
    mabowen01,
    lilincheng,
    dingyu01\}@corp.netease.com, xin.yu@uts.edu.au
}

\usepackage{bibentry}
\begin{document}

\maketitle

\begin{abstract}
In this work, we propose a semantic flow-guided two-stage framework for shape-aware face swapping, namely FlowFace.
Unlike most previous methods that focus on transferring the source inner facial features but neglect facial contours, our FlowFace can transfer both of them to a target face, thus leading to more realistic face swapping.
Concretely, our FlowFace consists of a face reshaping network and a face swapping network. 
The face reshaping network addresses the shape outline differences between the source and target faces. 
It first estimates a semantic flow (\emph{i.e.}, face shape differences) between the source and the target face, and then explicitly warps the target face shape with the estimated semantic flow. 
After reshaping, the face swapping network generates inner facial features that exhibit the identity of the source face. 
We employ a pre-trained face masked autoencoder (MAE) to extract facial features from both the source face and the target face. 
In contrast to previous methods that use identity embedding to preserve identity information, the features extracted by our encoder can better capture facial appearances and identity information. Then, we develop a cross-attention fusion module to adaptively fuse inner facial features from the source face with the target facial attributes, thus leading to better identity preservation.
Extensive quantitative and qualitative experiments on in-the-wild faces demonstrate that our FlowFace outperforms the state-of-the-art significantly.
\end{abstract}

\section{Introduction}
Face swapping refers to transferring the identity information of a source face to a target face while maintaining the attributes (e.g., expression, pose, hair, lighting, and background) of the target. 
It has attracted many interests due to its wide applications, such as portrait reenactment, film production, and virtual reality. 

Recent works~\cite{li2019faceshifter,chen2020simswap,xu2021facecontroller,li2021faceinpainter} have made great efforts to achieve promising face swapping results. 
However, these methods often focus on inner facial feature transferring but neglect facial contour reshaping.
We observe that facial contours also carry the identity information of a person, but few efforts~\cite{wang2021hififace} have been made on facial contours transferring. 
Facial shape transferring is still a challenge for authentic face swapping.

To solve the shape transferring problem, we propose a semantic flow-guided two-stage framework, dubbed FlowFace. Unlike existing methods, FlowFace is a shape-aware face swapping network. 
In a nutshell, we first present a face reshaping network to warp the target face referring to the source face shape at the first stage. Then, based on the reshaped facial contour, we further employ a face swapping network to transfer the inner facial features to the reshaped target face.

Our face reshaping network addresses the shape outline discrepancy between the source face and the target face.
Specifically, we use a 3D face reconstruction model (\emph{i.e.}, 3DMM \cite{blanz19993DMM}) to obtain shape coefficients of the source and target faces and then project the obtained 3D shapes to 2D facial landmarks. 
To accurately warp the target face, we need to obtain dense motion between the source and the target faces. Subsequently, we design a semantic guided generator to transform the sparse 2D facial landmarks into the dense flow.
The estimated flow, called semantic flow, will be exploited to warp the target face shape explicitly in a pixel-wise manner. In addition, we propose a semantic-guided discriminator to enforce our face reshaping network to produce accurate semantic flow.

After reshaping the target face, we introduce a face swapping network for transferring the inner facial features of the source face to the target ones. 
Prior works usually use a face recognition model to extract the identity embedding of the source face and then transfer it to the target face. 
We argue this would lose some personalized appearances during transferring because the identity embedding is often trained under discriminative tasks and thus may ignore intra-class variations~\cite{kim2022smooth}.
Thus, we opt to employ a pre-trained masked autoencoder (MAE)~\cite{he2022mae} to extract facial features that better capture facial appearances and identity information. 
Moreover, unlike prior arts that widely employ AdaIN~\cite{liu2017adaptive} to infuse the source identity embedding to the target face, we develop a cross-attention fusion module to adaptively fuse the source and target features.
In doing so, we achieve better face swapping performance.

Extensive quantitative and qualitative experiments validate the effectiveness of our FlowFace on in-the-wild faces and our FlowFace outperforms the state-of-the-art.
Overall, our contributions are summerized as follows:
\begin{itemize}
    \item We propose a two-stage framework for shape-aware face swapping, namely FlowFace. It can effectively transfer both the inner facial features and the facial outline to a target face, thus achieving authentic face swapping results.
    \item We design a semantic flow-guided face reshaping network and validate its effectiveness in transferring the source face shapes to the target ones. The reshaped target faces are more similar to the source faces in terms of face contours.
    \item We design a pre-trained face masked autoencoder based face swapping network. The encoder captures not only identity information but also facial appearance, thus allowing us to transfer richer information from the source face to the target and achieve identity similarity.
    \item We design a cross-attention fusion module to adaptively fuse the source and target features. To the best of our knowledge, we are the first to perform face swapping in the latent space of the pre-trained masked autoencoder.
\end{itemize}
%
\section{Related Work}
 
The previous face swapping methods can be classified as the target attribute-guided and source identity-guided methods.

\textbf{Target attribute-guided methods} edit the source face first and then blend it to the target background. 
Early methods~\cite{bitouk2008face,chen2019face,lin2012face} directly warp the source face according to the target facial landmarks, thus failing to address large posture differences and expression differences. 
3DMM-based methods~\cite{blanz2004exchanging,thies2016face2face,faceswap,nirkin2018face} swap faces by 3D-fitting and re-rendering. However, these methods often cannot handle skin color or lighting differences and suffer from poor fidelity. Later, GAN-based methods improve the fidelity of the generated faces. Deepfakes~\cite{deepfakes} transfers the target attributes to the source face by an encoder-decoder structure while being constrained by two specific identities.
FSGAN\cite{nirkin2019fsgan} employs the target facial landmarks to animate the source face and proposes a blending network to fuse the generated source face to the target background. However, it fails to tackle drastic skin color differences. 
Later, AOT~\cite{aot2020neurips} focuses on swapping faces with large differences in skin color and lighting by formulating appearance mapping as an optimal transport problem.
These methods always need a facial mask to blend the generated face with the target background. However, the mask-guided blending restricts the face shape change.

\textbf{Source identity-guided methods} usually adopt the identity embedding or the latent representation of StyleGAN2~\cite{karras2020analyzing} to represent the source identity and inject in into the target face. FaceShifter~\cite{li2019faceshifter} designs an adaptive attentional denormalization generator to integrate the source identity embedding and the target features. SimSwap~\cite{chen2020simswap} introduces a weak feature matching loss to help preserve the target attributes.  MegaFS~\cite{zhu2021one}, RAFSwap~\cite{xu2022region} and HighRes~\cite{xu2022high} utilize the pre-trained StyleGAN2 to swap faces and can achieve high-resolution face swapping.   FaceController~\cite{xu2021facecontroller} exploits the identity embedding with 3D priors to represent the source identity and design a unified framework for identity swapping and attribute editing. InfoSwap~\cite{gao2021information} leverages the information bottleneck principle to disentangle the identity and identity-irrelevant information. FaceInpainter~\cite{li2021faceinpainter} also utilizes the identity embedding with 3D priors to implement controllable face in-painting under heterogeneous domains. Smooth-Swap~\cite{kim2022smooth} builds smooth identity embedding that makes the training of face swapping fast and stable.
 
However, most of these methods neglect the facial outlines during face swapping. Recently, HifiFace~\cite{wang2021hififace} can control the face shape using a 3D shape-aware identity. However, it injects the shape representation into the latent feature space, making it hard for the model to correctly decode the face shape.
Moreover, these methods always need a pre-trained face recognition model during the inference time, which is not friendly to deployment.

\section{Proposed Method}

\begin{figure*}[t]
\centering
\includegraphics[width=1.0\textwidth]{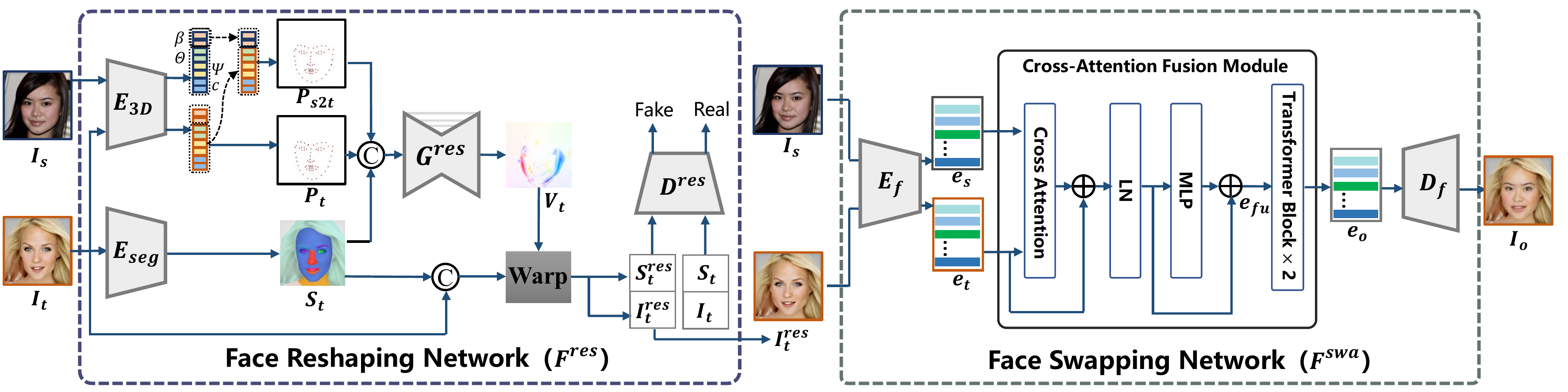}  
\caption{Overview of our two-stage FlowFace. In the first stage, the face reshaping network ($F^{res}$) transfers the shape of the source face $I_s$ to the target face $I_t$ by warping $I_t$ explicitly with an estimated semantic flow $V_t$. In the second stage, the face swapping network ($F^{swa}$) generates the inner facial details by manipulating the latent face representation $e_s$ and $e_t$ using our designed cross-attention fusion module. It should be noted that \textcircled{c} in the figure represents the concatenation operation.}
\label{fig:arch}
\end{figure*}

The face swapping task aims to generate a face with the identity of the source face and the attributes of the target face. This paper proposes a semantic flow-guided two-stage framework for shape-aware face swapping, namely FlowFace. As shown in Figure~\ref{fig:arch}, FlowFace consists of a face reshaping network $F^{res}$ and a face swapping network $F^{swa}$. Let $I_{s}$ be the source face and $I_{t}$ be the target face. $F^{res}$ first transfers the shape of $I_s$ to the target face $I_t$. The reshaped image is denoted as $I^{res}_{t}$. Then $F^{swa}$ generates the inner face of $I^{res}_{t}$ and outputs the result image $I_{o}$.

\subsection{Face Reshaping Network}
We design the face reshaping network, $F^{res}$, to address the shape discrepancy between the source and target faces. It warps the target face shape explicitly pixel-wise with an estimated semantic flow. To achieve this goal, $F^{res}$ requires a face shape representation that models the shape differences between the source and target faces. Then it estimates a semantic flow according to the above shape differences. Finally, the semantic flow is used to warp the target face shape.

\subsubsection{Face Shape Representation.}\label{para:face_shape_representation} 
Since our face reshaping network needs to warp the face shape pixel-wisely, we choose the explicit facial landmarks as the shape representation. We use a 3D face reconstruction model to obtain facial landmarks. As shown in Figure~\ref{fig:arch}, the 3D face reconstruction model $E_{3D}$ extracts 3D coefficients of the source and target: 
\begin{equation}
\begin{split}
(\beta_*, \theta_*, \psi_*, c_*)=E_{3D}(I_*),
\end{split}
\end{equation}
where $\beta_*, \theta_*, \psi_*, c_*$ are the FLAME coefficients~\cite{FLAME:SiggraphAsia2017} representing the face shape, pose, expression, and camera, respectively. $*$ is $s$ or $t$, representing the source or the target, respectively. With these coefficients, the target face can be modeled as:
\begin{equation}
M_t(\beta_t, \theta_t, \psi_t)=W\left(T_P(\beta_t, \theta_t, \psi_t), \mathbf{J}(\beta_t), \theta_t, \mathcal{W}\right),
\end{equation}
where $M_t$ represents the 3D face mesh of the target face. $W$ is a linear blend skinning (LBS) function that is applied to rotate the vertices of $T_P$ around joint $J$. $\mathcal{W}$ is the blend weights. $T_P$ denotes the template mesh $\overline{T}$ with shape, pose, and expression offsets~\cite{FLAME:SiggraphAsia2017}.

Then, we reconstruct the source face similarly, except that the source pose and expression coefficients are replaced with the target ones. The obtained 3D face mesh is denoted as $M_{s2t}$. Finally, we sample 3D facial landmarks from $M_t$ and $M_{s2t}$ and project these 3D points to 2D facial landmarks with the target camera parameter $c_t$:
\begin{equation}
\begin{split}
    P_{t} =s \Pi\left(M_{t}^{i}\right)+t, \\
    P_{s2t} =s \Pi\left(M_{s2t}^{i}\right)+t,
\end{split}
\end{equation}
where $M_{*}^{i}$ is a vertex in $M_{*}$, $\Pi$ is an orthographic 3D-2D projection matrix, and $s$ and $t$ are parameters in $c_t$, indicating isotropic scale and 2D translation. $P_{*}$ denotes the 2D facial landmarks. It should be noted that we only use the landmarks at the facial contours as the shape representation since inner facial landmarks contain identity information that may influence the reshaping result. 

\subsubsection{Semantic Flow Estimation.}\label{para:se_flow_estimation}

The relative displacement between $P_t$ and $P_{s2t}$ only describes sparse movement.
To accurately warp the target face, we need to obtain dense motion between the source and the target faces.
Therefore, we propose the semantic flow, which models the semantic correspondences between two faces, to achieve pixel-wised movement. We design a semantic guided generator $G^{res}$ to estimate the semantic flow. Specifically, $G^{res}$ requires three inputs: $P_{s2t}$, $P_t$ and $S_t$, where $P_{s2t}$ and $P_t$ are the 2D facial landmarks obtained above.
$S_t$ is the target face segmentation map that complements the semantic information lost in facial landmarks. The output of $G^{res}$ is the estimated semantic flow $V_t$, the formulation is:
\begin{equation}
V_t = G^{res}(P_{s2t}, P_t, S_t).
\end{equation}

Then, a warping module is introduced to generate the warped faces using $V_t$. We find that an inaccurate flow is likely to produce unnatural images, and therefore, we design a semantic guided discriminator $D^{res}$ that ensures $G^{res}$ to produce a more accurate flow. Specifically, the warping operation is conducted on both $I_t$ and $S_t$:
\begin{equation}
(I_t^{res},S_t^{res}) = F(V_t, I_t, S_t),
\end{equation}
where $F$ is the warping function in the warping module. We feed the concatenation of the warped face $I_t^{res}$ and the warped segmentation map $S_t^{res}$ to $D^{res}$. Thus, $D^{res}$ is able to discriminate whether the input is real or fake from the semantic level and the image level. It should be noted that $S_t^{res}$ and $D^{res}$ are only used during training.


\subsubsection{Training Loss.}
We employ three loss functions for $F^{res}$: 
\begin{equation}
\mathcal{L}^{res} = 
\mathcal{L}_{adv} + \lambda_{rec}\mathcal{L}_{rec} + \lambda_{ldmk}\mathcal{L}_{ldmk},
\label{res_loss_full}
\end{equation}
where $\lambda_{ldmk}$ and $\lambda_{rec}$ are hyperparameters for each term. In our experiments, we set $\lambda_{ldmk}$=800 and $\lambda_{rec}$=10.

\textbf{Adversarial Loss.}
To make the resultant images more realistic, we adopt the hinge version adversarial loss\cite{lim2017geometric} for training, denoted by $L_{adv}$:
\begin{equation}
\mathcal{L}_{adv}=-\mathbb{E}[D^{res}([I_t^{res},S_t^{res}])],
\end{equation}
where $D^{res}$ is the discriminator which is trained with:
\begin{equation}
\begin{split}
\mathcal{L}_{D}=\mathbb{E}[\max (0,1-D([I_t,S_t]))]\\+\mathbb{E}[\max (0,1+D([I_t^{res},S_t^{res}])] .
\end{split}
\end{equation}

\textbf{Reconstruction Loss.}
Since there is no ground-truth for face reshaping results, we enforce $I_s = I_t$ with a certain probability when training $G^{res}$. Then the face reshaping task becomes a reconstruction task, and we introduce a pixel-wise reconstruction loss:
\begin{equation}
\mathcal{L}_{rec}=\left\|{I_t^{res} - I_t}\right\|_2 \label{loss_rec},
\end{equation}
where $\left\|*\right\|_2$ denotes the euclidean distance.

\textbf{Landmark Loss.} Since there is not pixel-wised ground truth for $I_t^{res}$, we exploit the 2D facial landmarks $P_{s2t}$ to constrain the shape of $I_t^{res}$. Specifically, we first use a pre-trained facial landmark detector~\cite{sun2019high} to predict the the facial landmarks of $I_t^{res}$, denoted as $P_{t}^{res}$. Then the loss is computed as:
\begin{equation}
\mathcal{L}_{ldmk}=\left\|P_{t}^{res} - P_{s2t}\right\|_{2}.\
\end{equation}


At this point, our designed face reshaping network is able to transfer the face shape of the source to the target face. However, the inner facial features are still unchanged.

\subsection{Face Swapping Network}
The face swapping network $F^{swa}$ is used to generate the inner face of $I_t^{res}$ ($I_t$). As shown in Figure~\ref{fig:arch}, we first utilize a shared face encoder $E_{f}$ to map both $I_s$ and $I_t^{res}$ into patch embeddings $e_s$ nad $e_t$. Then a cross-attention fusion module is designed to adaptively fuse the identity information of the source face and the attribute information of the target. Finally, the facial decoder, fed with the manipulated embeddings $e_o$, outputs the final face swapping result $I_o$.

\subsubsection{Shared Face Encoder.}\label{para:encoder}
Most previous face swapping methods map the source face into an ID embedding with a pre-trained face recognition model and extract the target face attributes with another face encoder. 
However, we argue that using two different encoders is unnecessary and even makes deploying more complex. Moreover, the ID embedding is trained on purely discriminative tasks and may lose some personalized appearances during transferring.

Therefore, we employ a shared encoder to project both the source face and the target face into a common latent representation. 
The encoder is designed following MAE~\cite{he2022mae} and pre-trained on a large-scale face dataset using the masked training strategy. 
Compared to the compact latent code of StyleGAN2~\cite{karras2020analyzing} and the identity embedding, the latent space of MAE can better capture facial appearances and identity information, because masked training requires reconstructing masked image patches from visible neighboring patches, thus ensuring each patch embedding contains rich topology and semantic information. 

Based on the pre-trained encoder $E_f$, we can project a facial image $I_*$ into a latent representation, also known as patch embeddings:
\begin{equation}
e_* = E_f(I_*),
\end{equation}
where $e_* \in R^{N*L}$. $N$ and $L$ denote the number of patches and the dimension of each embedding, respectively. 

\subsubsection{Cross-Attention Fusion Module.}\label{para:cross_module}
The shared face encoder projects the source face and the target face into a representative latent space. The subsequent operation is to fuse the source identity information with the target attribute in this latent space. 
Intuitively, identity information should be transferred between related patches (\emph{e.g.}, nose to nose, etc.). Therefore, we design a cross-attention fusion module (CAFM) to adaptively aggregate identity information from the source and fuse it into the target.

As shown in Figure~\ref{fig:arch}, our CAFM consists of a cross-attention block and two standard transformer blocks~\cite{dosovitskiy2020vit}.
Given the source patch embeddings $e_{s}$ and the target patch embeddings $e_{t}$, we first compute $Q,K,V$ for each patch embedding in $e_{s}$ and $e_{t}$. Then the cross attention is computed by:
\begin{equation}
\operatorname{CA}(Q_t, K_s)=\operatorname{softmax}\left(\frac{Q_t K_s^{T}}{\sqrt{d_{k}}}\right),
\end{equation}
 where $\operatorname{CA}$ represents Cross Attention, $Q_*, K_*, V_*$ are predicted by attention heads, and $d_k$ is the dimension of $K_*$. 
 The cross attention describes the relation between each target patch and the source patches. 
 Next, the source identity information is aggregated based on the computed $\operatorname{CA}$ and fused to the target values via addition:
\begin{equation}
V_{fu}=\operatorname{CA} * V_s + V_t.
\end{equation}
 Then, $V_{fu}$ are normalized by a layer normalization (LN) and processed by multi-layer perceptrons (MLP). 
 The Cross Attention and MLP are along with skip connections. The fused embeddings $e_{fu}$ are further fed into two transformer blocks to obtain the final output $e_o$.

Finally, we utilize a a convolutional decoder to generate the final swapped face image $I_o$ from $e_o$. In contrast to the ViT decoder in MAE, we find the convolutional decoder achieves more realistic results.

\subsubsection{Training Loss.}
We employ six loss functions to train our face swapping network $F^{swa}$:
\begin{equation}
\begin{split}
\mathcal{L}^{swa} =  \mathcal{L}_{adv} + \lambda_{rec}\mathcal{L}_{rec} + \lambda_{id}\mathcal{L}_{id}  +  \lambda_{exp}\mathcal{L}_{exp} \\+ \lambda_{ldmk}\mathcal{L}_{ldmk} + \lambda_{perc}\mathcal{L}_{perc}
\end{split}, 
\label{swap_loss_full}
\end{equation}
where $\lambda_{rec}$, $\lambda_{id}$, $\lambda_{exp}$, $\lambda_{ldmk}$, $\lambda_{attr}$ are hyperparameters for each term. In our experiment, we set $\lambda_{rec}$=10, $\lambda_{id}$=5, $\lambda_{exp}$=10, $\lambda_{ldmk}$=5000 and $\lambda_{attr}$=2.

As in the face reshaping stage, the adversarial loss is used to make the resultant images more realistic, and the reconstruction loss between $I_o$ and $I_t^{res}$ is used for self-supervision since there is also no ground-truth for face swapping results. 

\textbf{Identity Loss.} The identity loss is used to improve the identity similarity between $I_s$ and $I_o$:
\begin{equation}
\mathcal{L}_{id} = 1- cos(E_{id}(I_o), E_{id}(I_s)), \label{loss_id}
\end{equation}
where $E_{id}$ denotes a face recognition model~\cite{deng2019arcface} and $cos$ denotes the cosine similarity. 

\textbf{Posture Loss.} We adopt the landmark loss to constrain the face posture during face swapping:
\begin{equation}
\mathcal{L}_{ldmk}=\left\|P_{t}^{res} - P_{o}\right\|_{2},\
\end{equation}
where $P_{o}$ represents the landmarks of $I_o$.

\textbf{Perceptual Loss.} Since high-level feature maps contain semantic information, we employ the feature maps from the last two convolutional layers of pre-trained VGG as the facial attribute representation. The loss is formulated as:
\begin{equation}
\mathcal{L}_{perc}=\left\|VGG(I_t^{res}) - VGG(I_o)\right\|_{2}.\
\end{equation}

\textbf{Expression Loss.} 
We utilize a novel fine-grained expression loss~\cite{zhang2021learning} that penalizes the $\mathcal{L}_2$ distance of two expression embeddings:
\begin{equation}
\mathcal{L}_{exp} = \left\|{E_{exp}(I_o) - E_{exp}(I_t)}\right\|_2. \label{loss_exp}
\end{equation}

\section{Experiments}

\begin{table}[t]
\setlength{\abovecaptionskip}{0cm}
\setlength{\belowcaptionskip}{-0.2cm}
\centering
\caption{Quantitative comparisons with state-of-the-art methods on FF++. "\dag" means the results are cited from their papers. 
}
\label{tab:quantitative_comparison}
\resizebox{0.45\textwidth}{!}{
\begin{tabular}{c|ccc|c|c|c}
\multirow{2}{*}{Methods} & \multicolumn{3}{c|}{ID Acc($\%$) $\uparrow$}                     & \multicolumn{1}{l|}{\multirow{2}{*}{Shape$\downarrow$}} & \multicolumn{1}{l|}{\multirow{2}{*}{Expr.$\downarrow$}} & \multicolumn{1}{l}{\multirow{2}{*}{Pose.$\downarrow$}} \\ \cline{2-4}
                         & CosFace  & SphereFace & Avg   & \multicolumn{1}{l|}{}                       & \multicolumn{1}{l|}{}                       & \multicolumn{1}{l}{}                       \\ \hline
Deepfakes              & 83.55     & 86.60      & 85.08 & 1.78                             & 0.54                             & 4.05                             \\
FaceSwap               & 70.95    & 76.77      & 73.86 & 1.85                             & 0.40                             & 2.21                             \\
FSGAN                  & 48.86     & 53.85      & 51.36 & 2.18                             & 0.27                             & 2.20                             \\
FaceShifter            & 97.38\dag     & 80.64      & 89.01 & 1.68                             & 0.33                             & 2.28                             \\
SimSwap                & 93.63      & 96.22      & 94.43 & 1.74                             & 0.26                             & \textbf{1.40}  \\ \hline
 $F^{res}$+SimSwap       & 94.31     & 96.82      & 95.56 & 1.67                             & 0.27                             & 2.27                             \\ 
HifiFace               & 98.48\dag     & 90.76      & 94.62 & 1.62                             & 0.30                             & 2.29                            
\\ \hline

$F^{swa}$          & \underline{99.18}      & \underline{98.23}      & \underline{98.70} & \underline{1.43}                             & \textbf{0.21}                             & \underline{1.99}                             \\
Ours & \textbf{99.26}      & \textbf{98.40}      & \textbf{98.83} & \textbf{1.17}                             & \underline{0.22}                             & 2.66             
\end{tabular}
}
\end{table}

\begin{figure*}[t]
\centering
\includegraphics[width=0.95\textwidth]{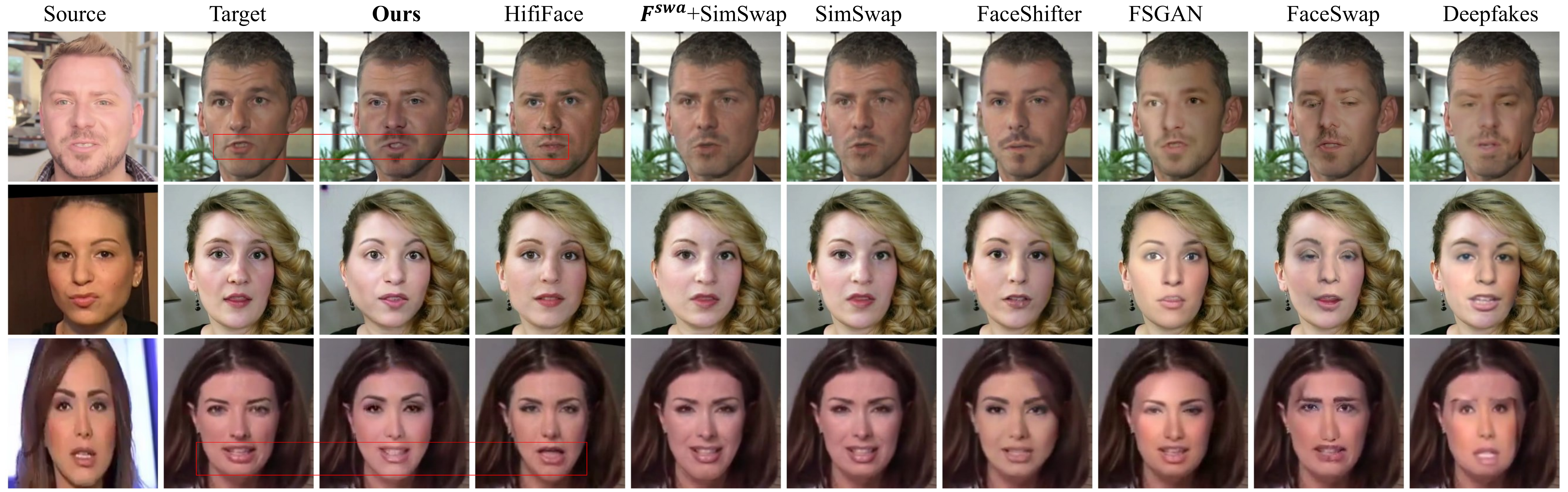}
\caption{Qualitative comparisons with Deepfakes, FaceSwap, FSGAN, FaceShifter, SimSwap and HifiFace on FF++. Our FlowFace outperforms the other methods significantly, especially in preserving face shapes, identities, and expressions. 
}
\label{fig:FF_comparison}
\end{figure*}

Our method is validated through qualitative and quantitative comparisons with state-of-the-art ones and a user study. Moreover, several ablation experiments are also reported to validate our design of FlowFace.

\subsection{Implementation Details}
\noindent{\textbf{Dataset.}} The training dataset is collected from three commonly-used face datasets: CelebA-HQ~\cite{karras2017progressive}, FFHQ~\cite{karras2019style}, and VGGFace2~\cite{cao2018vggface2}. Faces are aligned and cropped to $256\times256$.
Particularly, low-quality faces are removed to ensure high-quality training. The final dataset contains 350K face images, and 10K images are randomly sampled as the validation dataset. For the comparison experiments, we construct the test set by sampling FaceForensics++(FF++)~\cite{roessler2019faceforensicspp}, following~\cite{li2019faceshifter}. Specifically, FF++ consists of 1000 video clips, and the test set is collected by sampling ten frames from each clip of FF++, in a total of 10000 images.

\noindent{\textbf{Training.}} Our FlowFace is trained in a two-stage manner. Specifically, $F^{res}$ is first trained for 32K steps with a batch size of eight. As for $F^{swa}$, we first pre-trained the face encoder following the training strategy of MAE on our face dataset. Then we fix the encoder and train other components of $F^{swa}$ for 640K steps with a batch size of eight. 
We adopt Adam~\cite{kingma2014adam} optimizer with $\beta_1$=0 and $\beta_2$=0.99 and the learning rate is set to 0.0001. 
More details are in the supplementary materials and our codes will be made publicly available upon publication of the paper.

\noindent{\textbf{Metrics.}} The quantitative evaluations are performed in terms of four metrics: identity retrieval accuracy (ID Acc), shape error, expression error (Expr Error), and pose error.
We follow the same test protocol in~\cite{li2019faceshifter,wang2021hififace}. However, since some pre-trained models used in their testing are not available, we leverage different ones. For ID Acc, we employ two face recognition models, including CosFace~\cite{wang2018cosface} and SphereFace~\cite{liu2017sphereface}, to perform identity retrieval for a more comprehensive comparison. For expression error, we adopt a different expression embedding model~\cite{vemulapalli2019compact} to compute the euclidean distance of expression embeddings between the target and swapped faces.

\begin{figure*}[t]
\centering
\includegraphics[width=1.0\textwidth]{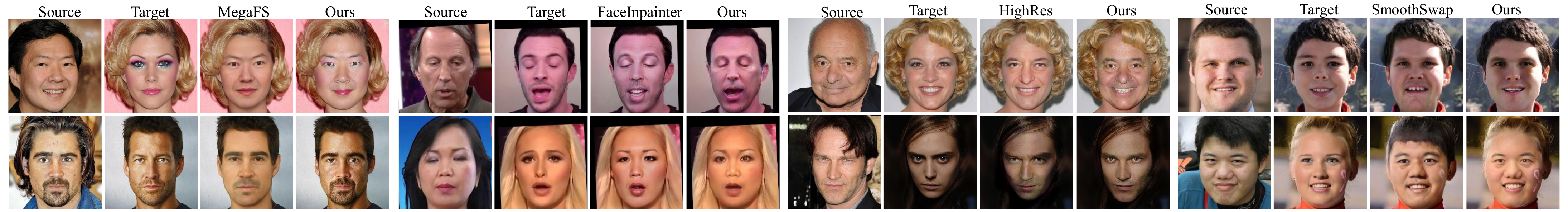}
\caption{Qualitative comparisons with more methods including  MegaFS~\cite{zhu2021one}, FaceInpainter~\cite{li2021faceinpainter}, HighRes~\cite{xu2022high} and SmoothSwap~\cite{kim2022smooth}. The shown images of the compared methods are cropped from their original papers or their released results.
}
\label{fig:additional_qualitative}
\end{figure*}

\subsection{Comparisons with the State-of-the-art}

\subsubsection{Quantitative Comparisons.} 
Our method is compared with six methods including Deepfakes~\cite{deepfakes}, FaceSwap~\cite{faceswap}, FSGAN~\cite{nirkin2019fsgan}, FaceShifter~\cite{li2019faceshifter}, SimSwap~\cite{chen2020simswap},  and HifiFace~\cite{wang2021hififace}.
For Deepfakes, FaceSwap, FaceShifter, and HifiFace, we use their released face swapping results of the sampled 10,000 images. For FSGAN and SimSwap, the face swapping results are generated with their released codes. 

Table~\ref{tab:quantitative_comparison} shows that our method achieves the best scores under most evaluation metrics, including ID Acc, shape error, and Expr Error. These results validate the superiority of our FlowFace. We obtain a slightly worse result than other methods for the pose error, which can be attributed to our FlowFace changing the face shape while the employed head pose estimator is sensitive to face shapes.

\begin{figure}[t]
\centering
\includegraphics[width=0.48\textwidth]{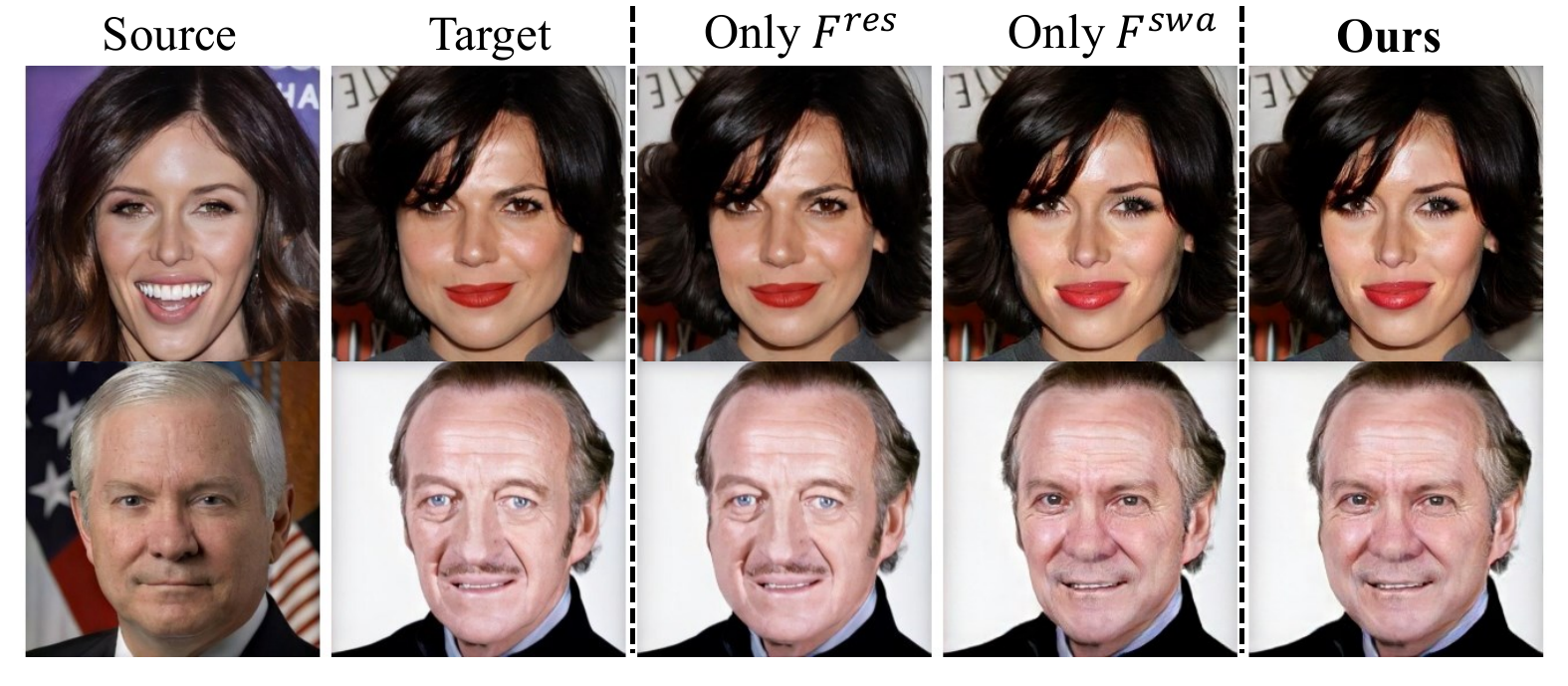}
\caption{
Qualitative ablation results of FlowFace.
}
\label{fig:face_swapping_result}
\end{figure}

\subsubsection{Qualitative Comparisons.} The qualitative comparisons are conducted on the same FF++ test set collected in the quantitative comparisons. As shown in Figure~\ref{fig:FF_comparison}, our FlowFace maintains the best face shape consistency. Note that most methods do nothing to transfer the face shape, so their resulting face shapes are similar to the target ones.

Although HifiFace is specifically designed to change the face shape, our method still obtains better results. As observed in Figure~\ref{fig:FF_comparison}, our generated face shapes are more similar to the source ones than HifiFace. Since HifiFace injects the shape representation into the latent feature space, it is harder to accurately decode the face shape from the latent feature than our explicit semantic flow.
Meanwhile, our method can better preserve the fine-grained target expressions (marked with red boxes in rows 1,3). 

\begin{table}[t]

\caption{Subjective comparisons with SimSwap and HifiFace on FF++.}
\centering
\label{tab:subjective_comparison}
\resizebox{0.40\textwidth}{!}{
\begin{tabular}{l|c|c|c|c}

Method      & Shape. ($\%$)$ \uparrow$  & ID. ($\%$)$ \uparrow$  & Exp. ($\%$)$ \uparrow$ & Realism ($\%$)$ \uparrow$ \\ \hline
SimSwap    & 20.67    & 27.78 & 34.44 & 14.67    \\ 
HifiFace & 35.11 & 34.67 & 30.45 & 41.78    \\ \hline
Ours        & \textbf{44.22} & \textbf{37.55} & \textbf{35.11} & \textbf{43.55} \\
\end{tabular}
}
\end{table}

\begin{figure*}[t]
\centering
\includegraphics[width=0.98\textwidth]{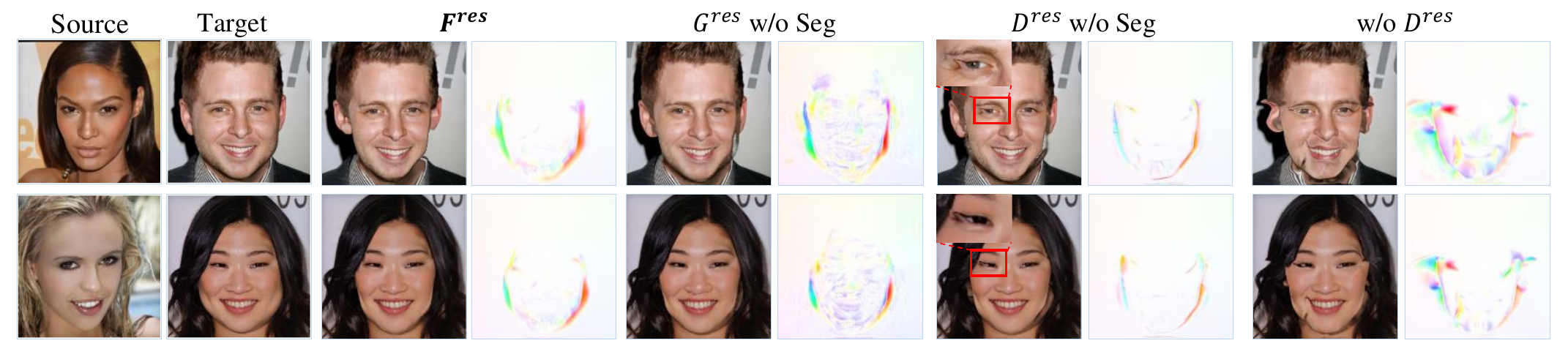}  
\caption{Qualitative ablation results of each component in $F^{res}$.}
\label{fig:ablation_first}
\end{figure*}

\begin{figure}[t]
\centering
\includegraphics[width=0.48\textwidth]{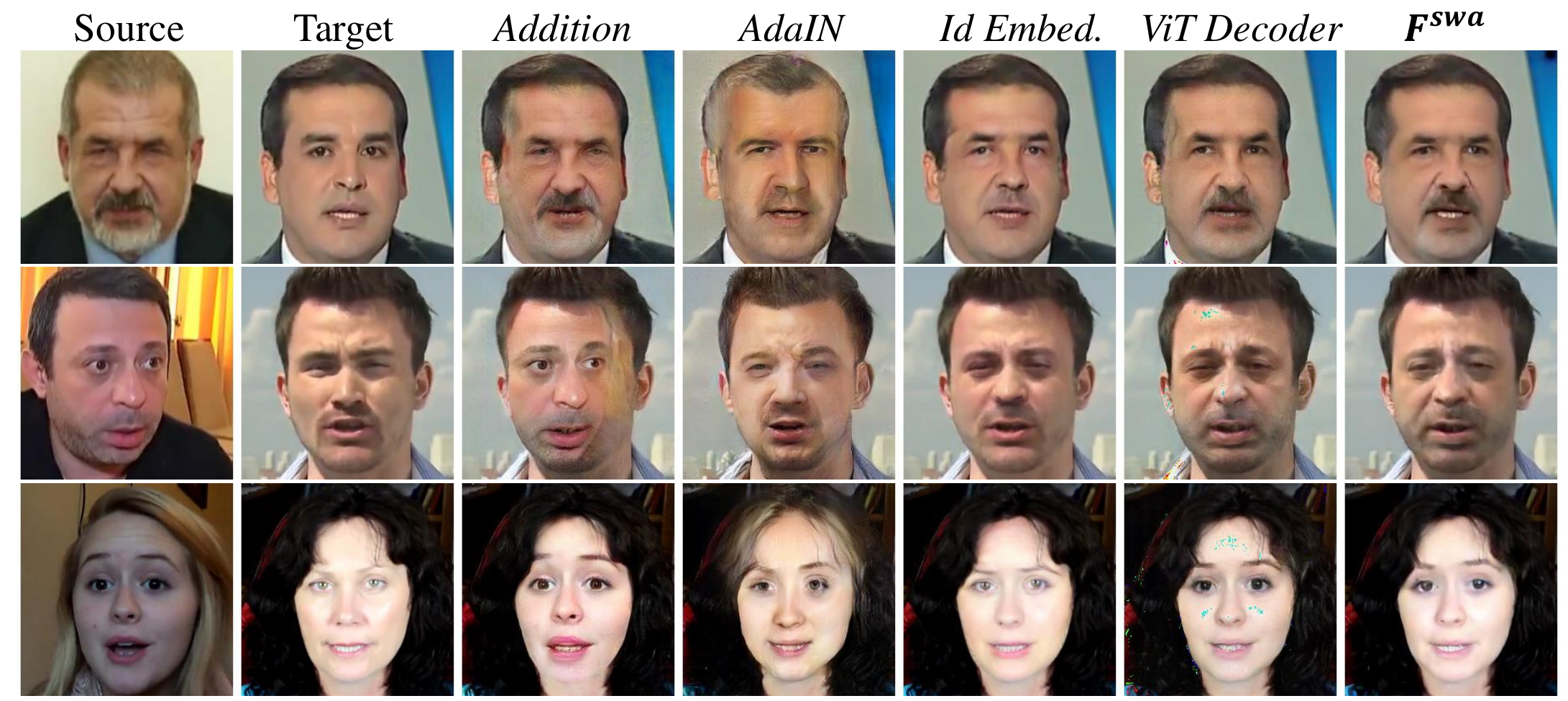}  
\caption{Qualitative ablation study of $F^{swa}$.}
\label{fig:ablation_second}
\end{figure}

\begin{figure}[t]
\centering
\includegraphics[width=0.45\textwidth]{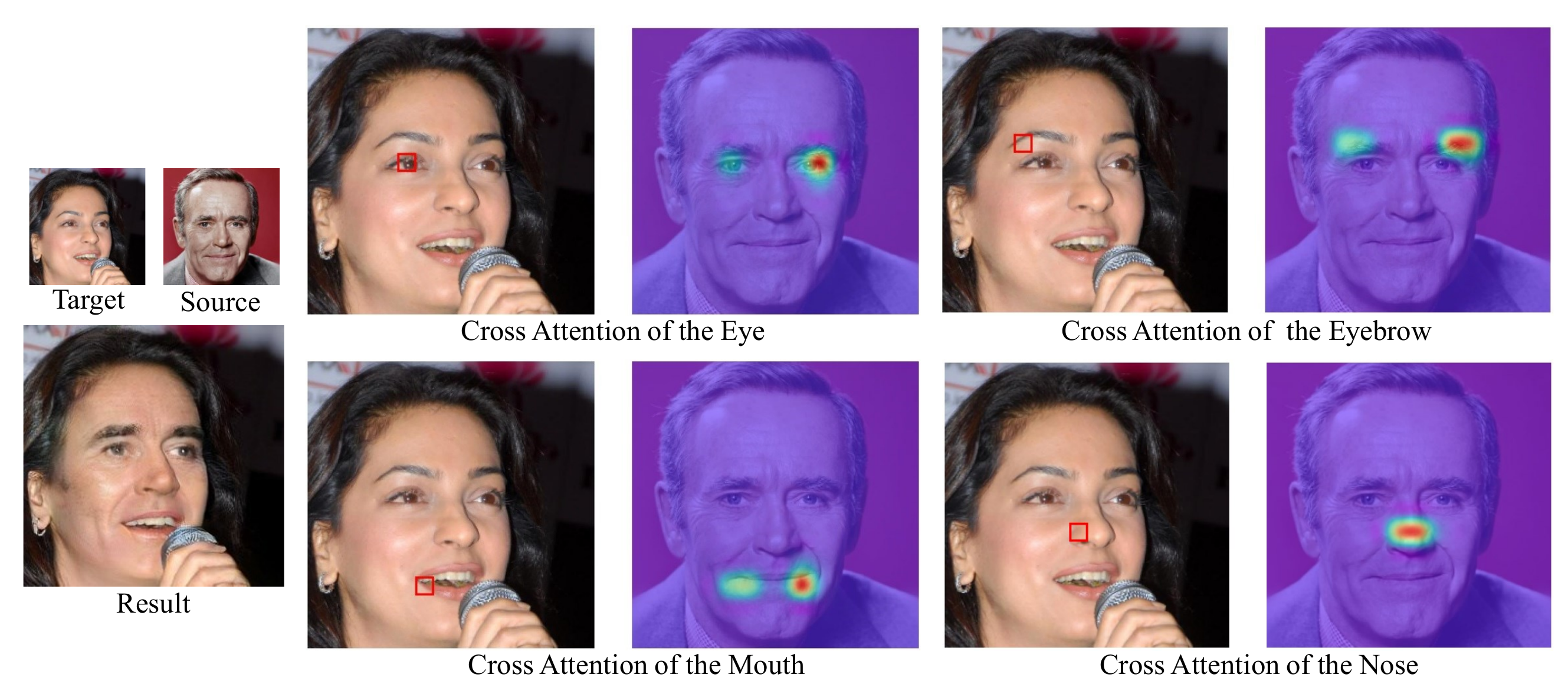}  
\caption{Visualize the cross-attention of different facial parts. For each part in the target, our CAFM can accurately focus on the corresponding parts in the source.
}
\label{fig:cross_att_vis}
\end{figure}

We further compare our methods with four more SOTA face swapping methods:~\cite{zhu2021one}, FaceInpainter~\cite{li2021faceinpainter}, HighRes~\cite{xu2022high} and SmoothSwap~\cite{kim2022smooth}. Among them, MegaFS and HighRes are based-on the latent space of StyleGAN2. As shown in Figure~\ref{fig:additional_qualitative}, our method can better transfer the shape of the source to the target than all other methods. Although SmoothSwap can change the face shape, it destroys the target attributes (\emph{e.g.}, hairstyle and hair color). Besides, our results are also more similar to the source face in terms of inner facial features (\emph{e.g.}, beard), validating that our face encoder can better capture facial appearances than the identity embedding or the latent code of StyleGAN2.
Moreover, our method also preserves the target attributes (\emph{e.g.}, skin color, lighting, and expression) better than other methods.

\subsubsection{User Study.}
To further validate our FlowFace, we conduct a subjective comparison with SimSwap and HifiFace, two SOTA methods that release their codes or results. Fifteen participants are instructed to choose the best result in terms of shape consistency, identity consistency, expression consistency, or image realism, involving comparisons of 30 swapped faces by three methods. Table~\ref{tab:subjective_comparison} shows that our method outperforms the two baselines in terms of all four metrics, validating the superiority of our method.

\subsection{Analysis of FlowFace}

Three ablation studies are conducted to validate our two-stage FlowFace framework and several components used in $F^{res}$ and $F^{swa}$, respectively.

\subsubsection{Ablation Study on FlowFace.}

We conduct ablation experiments to validate the design of our two-stage framework.
Figure~\ref{fig:face_swapping_result} shows the swapped images by only $F^{res}$, only $F^{swa}$ and the full model (FlowFace). 
It can be seen that $F^{res}$ transforms the face shape naturally according to the source, while $F^{swa}$ is good at capturing the identity of the source inner face and other facial attributes of the target. 
Benefiting from the strengths of both $F^{res}$ and $F^{swa}$, our FlowFace is able to create results with accurate identity and consistent facial attributes. Table~\ref{tab:quantitative_comparison} records the quantitative results, which further illustrates the effectiveness of our two-stage framework. 
The above observation validates the effectiveness of $F^{res}$ and confirms that face shapes are essential for identifying a person.

 To further validate our $F^{res}$, we plug it into the open-sourced SimSwap. As shown in Figure~\ref{fig:FF_comparison} and Table~\ref{tab:quantitative_comparison}, after reshaping by $F^{res}$, the face swapping result of SimSwap is more similar to the source face in terms of face contours. 
 The ID Acc. also rises from 93.63\% to 94.31\%. The results demonstrate the effectiveness of our $F^{res}$ and also reveal that the face shape carries the identity information, thus improving identity similarity.

\subsubsection{Ablation study on $F^{res}$.}

We first conduct an ablation experiment to validate our proposed semantic guided generator $G^{res}$. 
Specifically, we remove the semantic input $S_t$ of $G^{res}$ ($G^{res}$ w/o Seg). It can be seen from Figure~\ref{fig:ablation_first} that some inaccurate flow occurs in the generated face, which implies that only facial landmarks cannot guide $G^{res}$ to produce accurate dense flow due to the lack of semantic information. 
The results also demonstrate that the semantic information is beneficial for accurate flow estimation and validates the effectiveness of $G^{res}$.

Then, we conduct two ablation experiments to validate $D^{res}$: (1) removing the semantic inputs ($S_t$ and $S_t^{res}$) of $D^{res}$ ($D^{res}$ w/o Seg). 
Compared with $F^{res}$, the generated faces suffer from unnaturalness, like the eyes are stretched, as observed in Figure~\ref{fig:ablation_first}. 
It implies that structured information in the semantic inputs can provide more fine-grained discriminative signals, thus enforcing $G^{res}$ to produce a more accurate flow. 
(2) removing $D^{res}$ (w/o $D^{res}$). As observed in Figure~\ref{fig:ablation_first}, compared with $F^{res}$, there are many artifacts in the generated images, and the estimated flow also contains many noises. 
The above observation validates the effectiveness of our proposed $D^{res}$.

\begin{table}[t]

\setlength{\abovecaptionskip}{0cm}
\setlength{\belowcaptionskip}{-0.2cm}
\centering
\caption{Quantitative ablation study of $F^{swa}$ on FF++.}
\label{tab:ablation_swap}
\resizebox{0.45\textwidth}{!}{
\begin{tabular}{l|ccc|c|c}
\multirow{2}{*}{Methods} & \multicolumn{3}{c|}{ID Acc}                     & \multicolumn{1}{l|}{\multirow{2}{*}{Expr}} & \multicolumn{1}{l}{\multirow{2}{*}{Pose}} \\ \cline{2-4}
                         & CosFace        & SphereFace     & Avg            & \multicolumn{1}{l|}{}                       & \multicolumn{1}{l}{}                       \\ \hline
\textit{Addition}                 & \textbf{99.38}          & \textbf{99.44} & \textbf{99.41} & 0.43                                        &   4.90                                        \\
\textit{AdaIN}                 & 97.31          & 97.15 & 97.23 & 0.33                                        &   3.27                                        \\
\hline
\textit{Id Embed.}                   & 97.10          & 96.90          & 97.00          & \underline{0.22}                                  & \underline{2.10}                                          \\
\textit{Vit Decoder}                & 98.44          & 97.73          & 98.09          & 0.23                                        & 2.80                                         \\
$F^{swa}$                     & \underline{99.18} & \underline{98.23}    & \underline{98.71}    & \textbf{0.21}                               & \textbf{1.99}                                      
\end{tabular}
}
\end{table}

\subsubsection{Ablation study on $F^{swa}$.}
Three ablation experiments are conducted to evaluate the design of $F^{swa}$:
 
(1) Choices on CAFM, \textit{Addition} and \textit{AdaIN}. To verify the effectiveness of CAFM, we compare with two other methods: \textit{Addition} that directly adds the source values to the target values; \textit{AdaIN} that first averages source patch embeddings and then injects it into the target feature map using AdaIN residual blocks.
As shown in Figure~\ref{fig:ablation_second} and Table~\ref{tab:ablation_swap}, \textit{Addition} simply brings all information of the source face to the target face, thus leading to severe pose and expression mismatch. \textit{AdaIN} impacts the non-face parts (\emph{e.g.}, hair) due to its global modulation. In contrast, $F^{res}$ with CAFM obtains a high ID Acc and preserves the target attribute well, which proves that CAFM can accurately extract identity information from the source face and adaptively infuse it into the target counterpart. 

To further validate the effectiveness of our CAFM, we visualize the cross attention computed by CAFM. As shown in Figure~\ref{fig:cross_att_vis}, given a specific part (marked by red boxes) of the target face, CAFM accurately focuses on the corresponding parts of the source face, validating our CAFM can adaptively transfer the identity information from the source patches to corresponding target patches.

(2) Latent Representation vs. ID Embedding (\textit{ID Embed.}). 
To verify the superiority of using the latent representation of MAE, we train a new model which adopts the identity embedding as the identity representation and employs AdaIN as the injection method.
As can be seen from Figure~\ref{fig:ablation_second}, \textit{ID Embed.} misses some fine-grained face appearances, such as eyes color, beard. 
In contrast, $F^{swa}$ contains richer identity information and achieves higher ID Acc, as shown in Tab~\ref{tab:ablation_swap}.

(3) Convolutional Decoder vs. ViT Decoder (\textit{ViT Decoder}). 
We try two different decoders to find out the better one.
As shown in Figure~\ref{fig:ablation_second}, the results of \textit{ViT Decoder} contains a lot of artifacts. 
In contrast, \textit{Convolutional Decoder} achieves realistic results with high fidelity. 

\section{Conclusion}
This work proposes a semantic flow-guided two-stage framework, FlowFace, for shape-aware face swapping. 
In the first stage, the face reshaping network transfers the shape of the source face to the target face by warping the face pixel-wisely using semantic flow. In the second stage, we employ a pre-trained masked autoencoder to extract facial features that better capture facial appearances and identity information. Then, we design a cross-attention fusion module to better fuse the source and the target features, thus leading to better identity preservation. Extensive quantitative and qualitative experiments are conducted on in-the-wild faces, demonstrating that our FlowFace outperforms the state-of-the-art significantly. 

\section{Acknowledgments}
This work is supported by the 2022 Hangzhou Key Science and Technology Innovation Program (No. 2022AIZD0054), and the Key Research and Development Program of Zhejiang Province (No. 2022C01011), the ARC-Discovery grants (DP220100800) and ARC-DECRA (DE230100477).

\bibliography{egbib}

\newpage

\appendix
\section*{Supplementary Material}

\section{Implementation Details}
All models are implemented with PyTorch ~\cite{NEURIPS2019_9015}. We train these models in a two-stage manner. We first train the face reshaping network and then the face swapping network. For both stages, we adopt Adam~\cite{kingma2014adam} optimizer with $\beta_1$=0 and $\beta_2$=0.99 and the learning rate is set to 0.0001. For stable training, we apply spectral normalization~\cite{miyato2018spectral} to discriminators in both stages. All experiments are conducted on NVIDIA RTX A30 GPUs.

\subsubsection{Face Reshaping Network} 
\begin{figure}[t]
\centering
\includegraphics[width=0.48\textwidth]{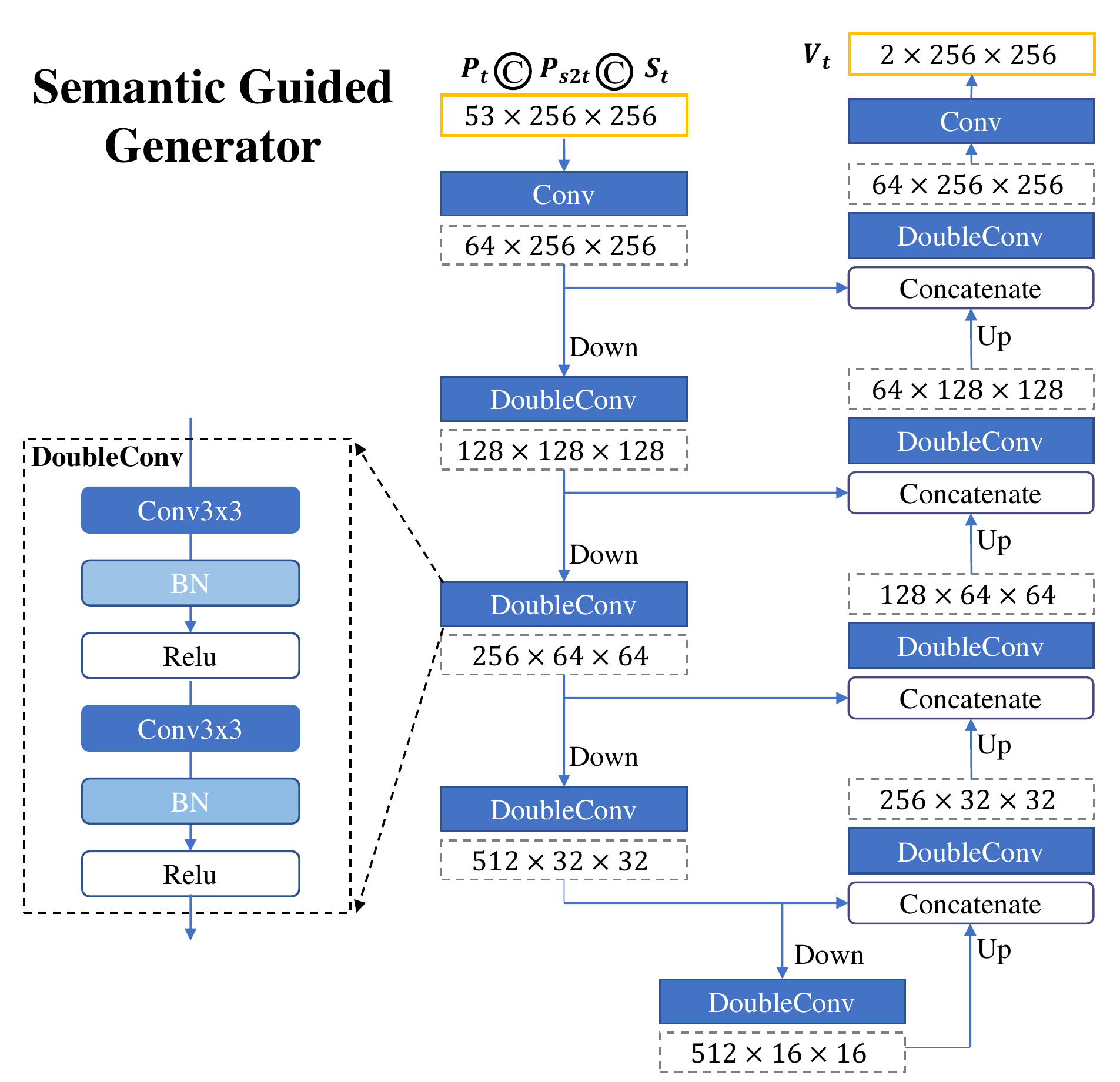}
\caption{
    The architecture of our semantic guided generator in the face swapping network. We adopt 17 landmarks to represent the face shape and these landmarks are transformed into heatmaps. The segmentation map is a tensor with 19 channels, representing 19 components in a face image. Therefore, the input tensor of the generator has 53 channels.
}
\label{fig:semantic_G}
\end{figure}

Our face reshaping network consists of a semantic guided generator and a semantic guided discriminator. As shown in Figure~\ref{fig:semantic_G}, we only show the detailed structure of the semantic guided generator, which is based on a U-net~\cite{ronneberger2015u}. We design the semantic guided discriminator by following PatchGAN~\cite{isola2017image}. Specifically, we change the input channel from 3 to 22, since the input of our semantic guided discriminator contains an additional face segmentation map, which has 19 channels.

\subsubsection{Face Swapping Network} 
\begin{figure*}[t]
\centering
\includegraphics[width=0.90\textwidth]{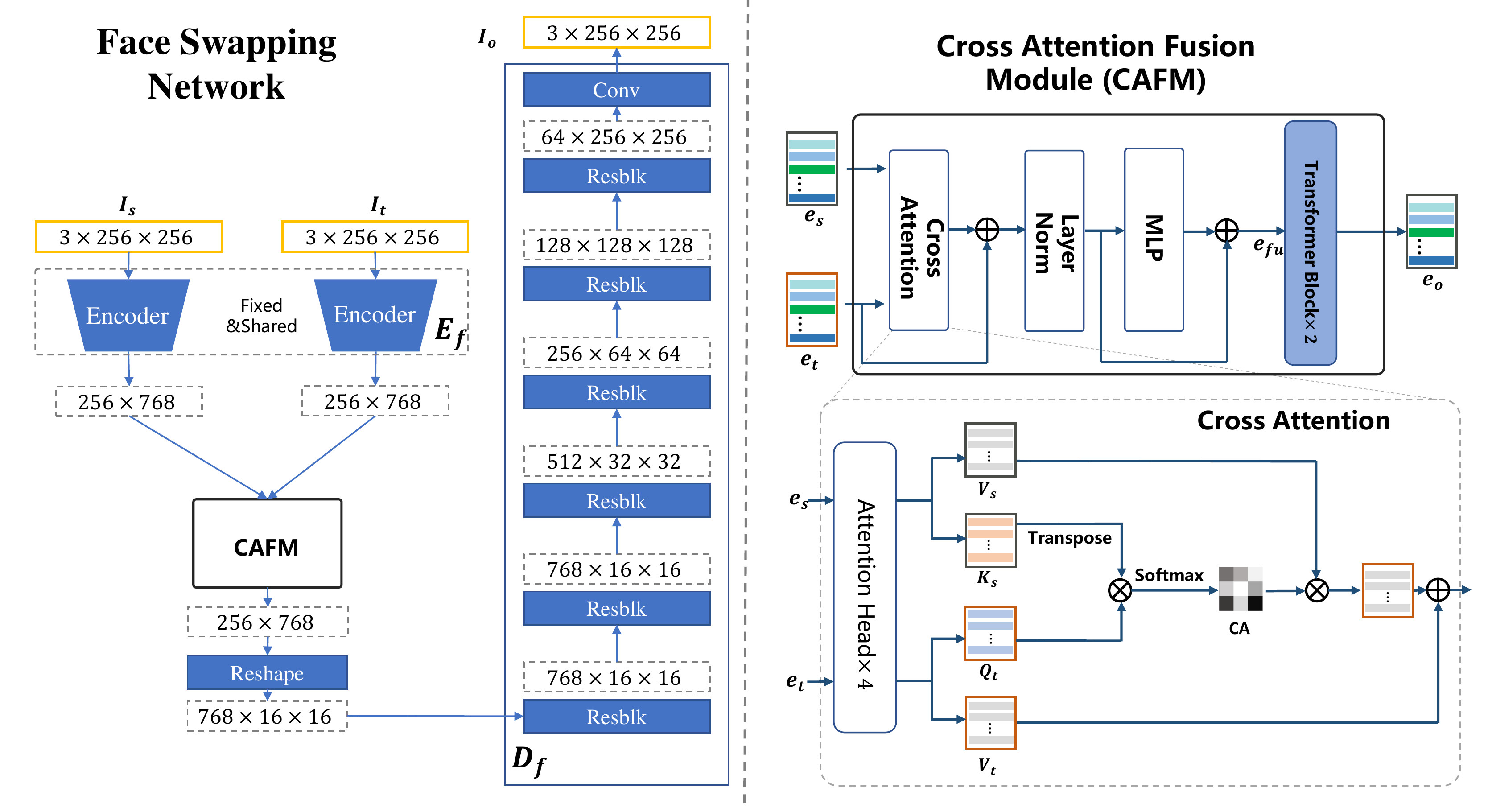}
\caption{
    The architecture of our face swapping network. We draw the details of each component in the face swapping network, including the facial encoder $E_f$, the mask decoder $D_m$, the facial decoder $D_f$ and the identity compound embedding encoder $E_c$.
}
\label{fig:swapping_G}
\end{figure*}

As shown in Figure~\ref{fig:swapping_G}, we draw the details of each component in the face swapping network, including the shared facial encoder $E_f$, the cross attention fusion module $\operatorname{CAFM}$ and the facial decoder $D_f$. The discriminator and the residual blocks are the same as StarGANv2~\cite{choi2020stargan}. The transformer blocks are standard as in ViT~\cite{dosovitskiy2020vit}.

\subsubsection{Face Reshaping Network Training.}
We only sample 10000 face images from our constructed dataset for training, and the 3DMM coefficients of these faces are extracted by the pre-trained DECA model\footnote{https://github.com/YadiraF/DECA}~\cite{feng2020learning} offline. 
During training, we first construct a batch size by randomly sampling eight source-target image pairs and their corresponding coefficients. The probability for $I_s = I_t$ is set to 10\%. 
Then, the facial landmarks $P_t$ and $P_{s2t}$ are computed online. The facial segmentation maps are also obtained online using the pre-trained face parsing model\footnote{https://github.com/zllrunning/face-parsing.PyTorch}.
For the landmark loss, we adopt HRNet\footnote{https://github.com/HRNet/HRNet-Facial-Landmark-Detection}~\cite{sun2019high}.
The face reshaping network is trained for 25 epochs.

\begin{figure*}[ht]
\centering
\includegraphics[width=0.90\textwidth]{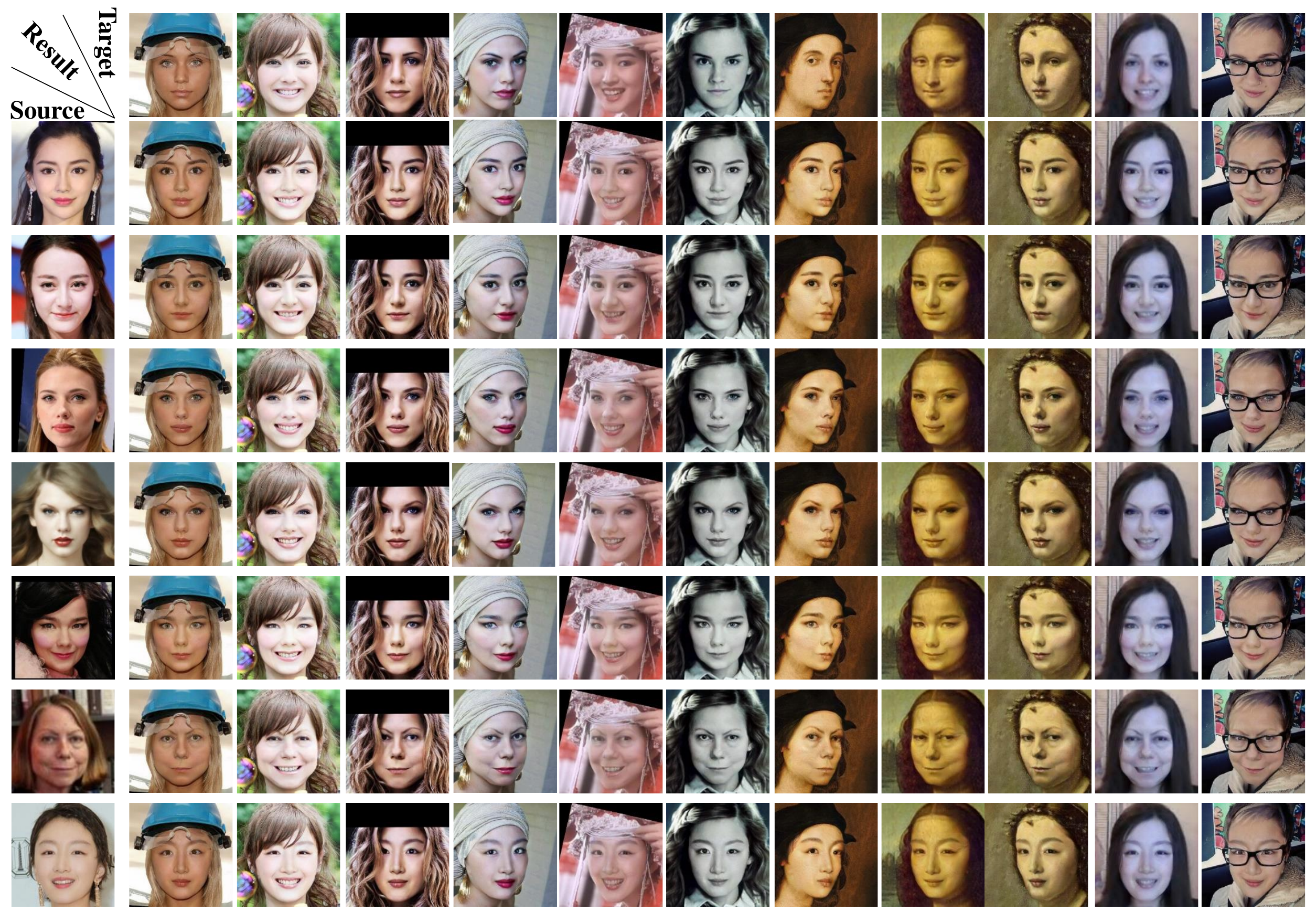}
\caption{
    More face swapping results of female faces. The images contain various postures, lighting conditions, expressions, skin colors, and makeup. Moreover, our FlowFace can also tackle artistic paintings. Please zoom in for more details.
}
\label{fig:more_results_female}
\end{figure*}
\begin{figure*}[t]
\centering
\includegraphics[width=0.95\textwidth]{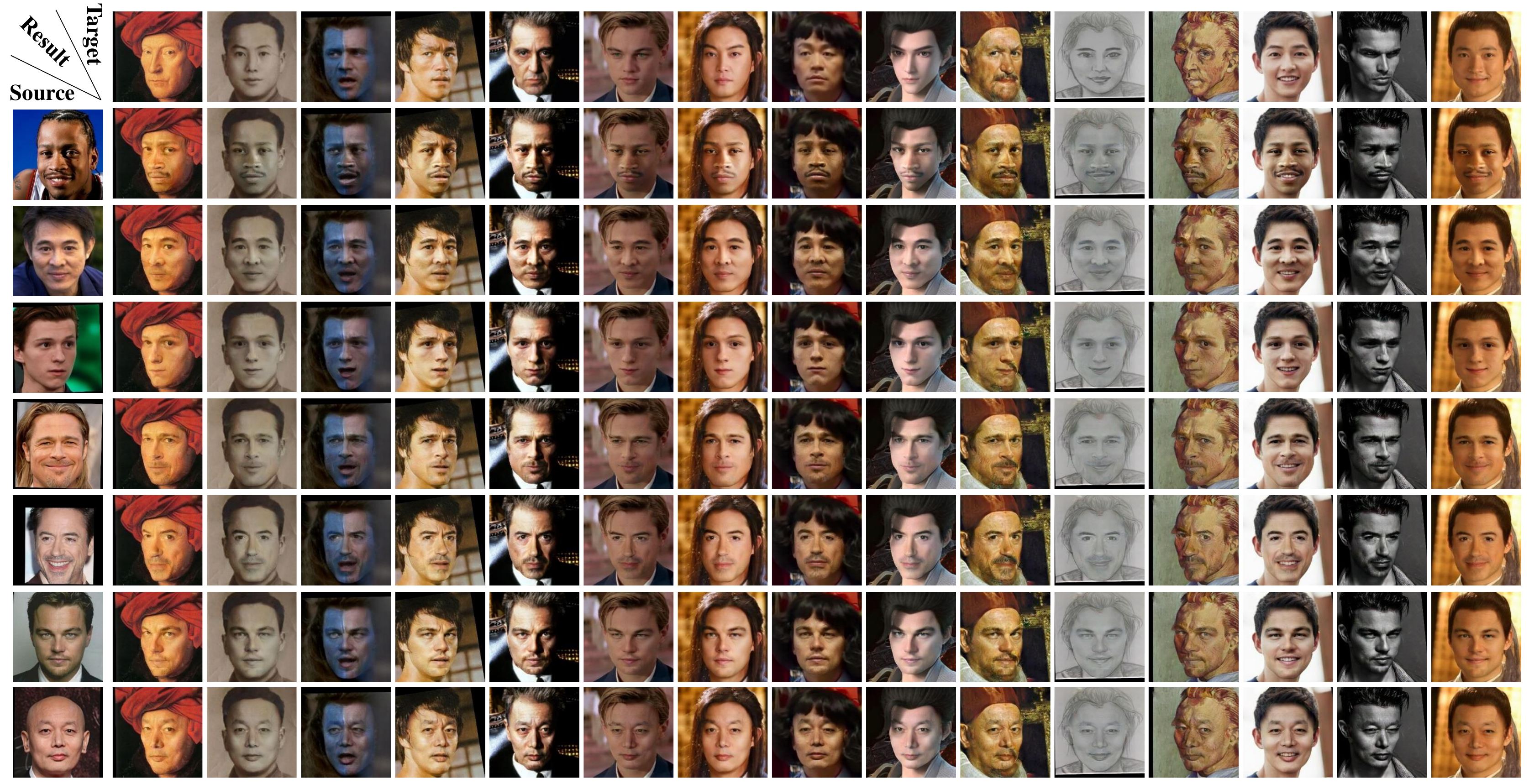}
\caption{
    More face swapping results of male faces. 
}
\label{fig:more_results_male}
\end{figure*}

\subsubsection{Face Swapping Network Training.}

We first train our face masked autoencoder, following the same settings in MAE\footnote{https://github.com/facebookresearch/mae}~\cite{he2022mae}. Then, we preserve the encoder part of MAE as our shared face encoder and fix it all the time during training. 
We train our face swapping network for 30 epochs on our constructed training set. The batch size is set to eight, and the probability for $I_s = I_t$ is set to 25\%.

\section{Additional Results}

Figure~\ref{fig:more_results_female} and Figure~\ref{fig:more_results_male} illustrate more face swapping results generated by our FlowFace. The images are captured under various conditions (e.g., postures, expressions, skin colors, lighting conditions, makeup, etc.), and the faces also come from various domains (e.g., human faces, cartoon faces, and art painting faces).

\begin{figure*}[ht]
\centering
\includegraphics[width=0.6\textwidth]{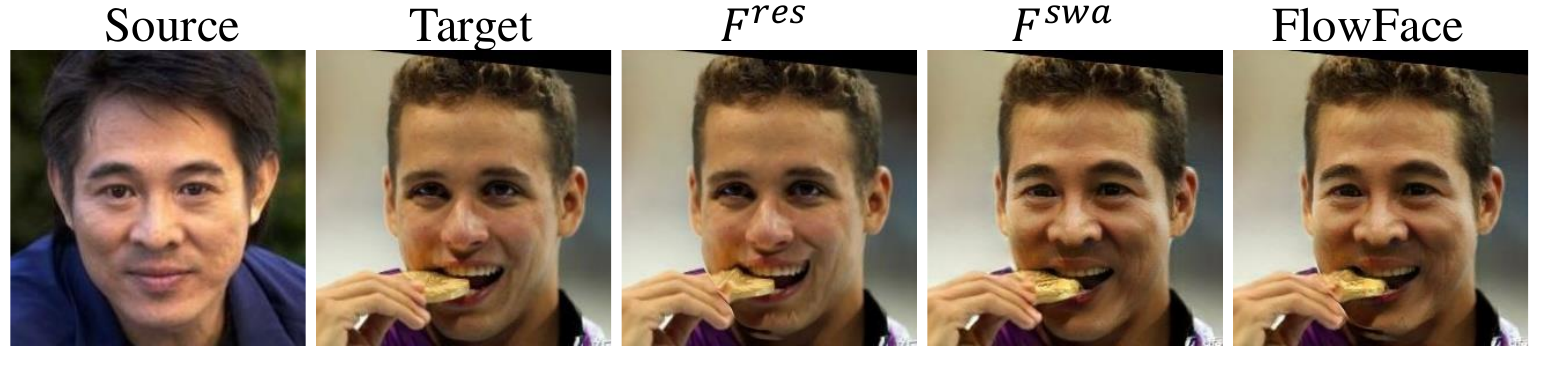}
\caption{
    Some failure cases of FlowFace. Occlusions may lead to inaccurate semantic flow in the face reshaping network. However, our face swapping network preserves the occlusion well.
}
\label{fig:failure_case}
\end{figure*}

\section{Limitations and Future Work}
Although our method achieves superior visual results to prior arts, it also shares some limitations. As shown in Figure~\ref{fig:failure_case}, our face reshaping network may be influenced by some occlusions, even though the face swapping network can well preserve the occlusions. However, this limitation can be addressed by the refinement network as in~\cite{li2019faceshifter} or an occlusion-specific attribute loss. Besides, the resultant video may suffer from flicker when our FlowFace is applied to video sequences frame by frame. This is because the reshaping network relies on 3D face reconstruction and facial landmarks, and neither of them can preserve the temporal consistency. However, we will address the temporal consistency problem in our future work by either a post-processing operation or a new temporal consistent framework.

\end{document}